\documentclass[lettersize,journal]{IEEEtran}
\usepackage{amsmath,amsfonts}
\usepackage{algorithmic}
\usepackage{algorithm}
\usepackage{array}
\usepackage[caption=false,font=normalsize,labelfont=sf,textfont=sf]{subfig}
\usepackage{textcomp}
\usepackage{stfloats}
\usepackage{url}
\usepackage{verbatim}
\usepackage{graphicx}
\usepackage{cite}
\usepackage{multirow}
\usepackage{bm}
\begin{document}

\title{OBBStacking: An Ensemble Method for Remote Sensing Object Detection}

\author{Haoning Lin, Changhao Sun and Yunpeng Liu
\thanks{Haoning Lin, Changhao Sun and Yunpeng Liu are with Key Laboratory of Opto-Electronic Information Processing, Chinese Academy of Sciences, Shenyang 110016, China, and also with Shenyang Institute of Automation, Chinese Academy of Sciences, Shenyang 110016, China, and also with Institutes for Robotics and Intelligent Manufacturing, Chinese Academy of Sciences, Shenyang 110169, China}
\thanks{Manuscript received xxxx, 2022; revised xxxx, 2022.}}

\markboth{Journal of \LaTeX\ Class Files,~Vol.~14, No.~8, August~2021}%
{Shell \MakeLowercase{\textit{et al.}}: A Sample Article Using IEEEtran.cls for IEEE Journals}

\IEEEpubid{0000--0000/00\$00.00~\copyright~2021 IEEE}

\maketitle

\begin{abstract}
    Ensemble methods are a reliable way to combine several models to achieve superior performance. However, research on the application of ensemble methods in the remote sensing object detection scenario is mostly overlooked. Two problems arise. First, one unique characteristic of remote sensing object detection is the Oriented Bounding Boxes (OBB) of the objects and the fusion of multiple OBBs requires further research attention. Second, the widely used deep learning object detectors provide a score for each detected object as an indicator of confidence, but how to use these indicators effectively in an ensemble method remains a problem. Trying to address these problems, this paper proposes OBBStacking, an ensemble method that is compatible with OBBs and combines the detection results in a learned fashion. This ensemble method helps take 1st place in the Challenge Track \textit{Fine-grained Object Recognition in High-Resolution Optical Images}, which was featured in \textit{2021 Gaofen Challenge on Automated High-Resolution Earth Observation Image Interpretation}. The experiments on DOTA dataset and FAIR1M dataset demonstrate the improved performance of OBBStacking and the features of OBBStacking are analyzed. Code will be available at \url{https://github.com/Haoning724/obbstacking}.
\end{abstract}

\begin{IEEEkeywords}
    Remote sensing, ensemble, object detection, stacking, oriented bounding box.
\end{IEEEkeywords}

\section{Introduction}
\IEEEPARstart{W}{ith} deep learning, researchers can design arbitrarily structured models as they see fit to a specific problem, which in turn leads to a wide range of off-the-shelf deep learning models. Ensemble methods are a reliable way to combine these models and achieve stronger performance. However, in the remote sensing object detection scenario, the potential of ensemble methods is rarely exploited.

Non-Maximum Suppression (NMS) \cite{felzenszwalb_object_nodate} is a widely used method to suppress redundant detection bounding boxes in a close neighborhood, by clustering the overlapped bounding boxes (BBs) and eliminating the non-confidence-maximum BBs in each cluster. Beyond its wide application in single object detectors, it can also be used as a simple ensemble method. However, NMS adopts an \textit{affirmative} voting strategy and thus assumes all of the detection results are true positives, and favors the models that vote for a detected object over those that vote against it.

Weighted Boxes Fusion (WBF) \cite{solovyev_weighted_2021} aims to alleviate the weakness of NMS, by taking into account all the confidence scores of the to-be-fused bounding boxes and assigning an average confidence score to the resulting bounding boxes. 

\begin{figure}[t]
    \centering
    \includegraphics[width=.4\textwidth]{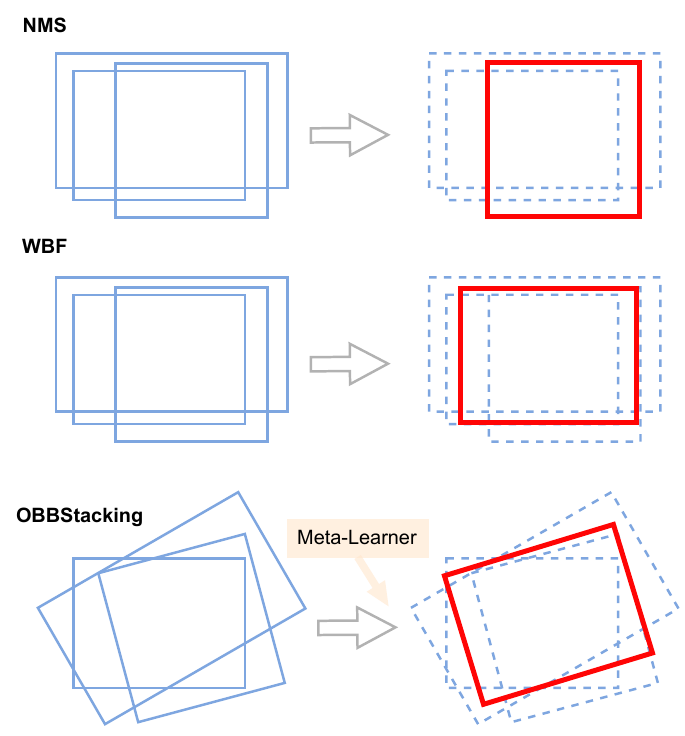}
    \caption{Ilustration of the bounding box fusion results of different ensemble methods. The blue rectangles are the bounding boxes fed into the methods, the red rectangles are the fused bounding boxes.}\label{fig:schematic}
\end{figure}

This method, however, leaves two problems unaddressed.

First, WBF treats the confidence scores from different models equally and takes the non-weighted mean value as the fused confidence score, disregarding three facts: 1. Some models may perform better than other models and their scores should have more weight. 2. Some models may share a similar neural network structure and produce similar results, so the ensembled result may bias towards a group of similar-structured models. 3. Deep learning models are poorly calibrated and different models will be overconfident to different extents, so a simple ensemble method may favor the more overconfident models.

\IEEEpubidadjcol
Second, WBF is only compatible with horizontal bounding boxes.

When deep learning was first introduced into the remote sensing object detection problem, the position of a detected object was initially encoded in the same format as those in the other scenarios, i.e. a non-oriented rectangular bounding box with its sides always horizontal to either one of the side of the image coordinate grids. This format soon posed a problem. Due to the high altitude viewpoint and the steep viewing angle of the remote sensing images, the presented objects can have arbitrary orientations. Some types of objects, such as large ships, buses, buildings, and airport runways, have a large length-to-width ratio and are poorly represented by horizontal bounding boxes, especially when the objects are at a roughly $\pm 45 ^{\circ}$ angle to the image axes. 

Oriented Bounding Box (OBB) was proposed to address this problem. OBB keeps the rectangular form but obtains \textit{orientation} as a new degree of freedom (DoF), the other existing DoFs being the position of its center, length, and width. OBB introduces finer labels to the objects in the remote sensing images and a better data format for the detection accuracy criteria. However, the existing ensemble methods are not compatible with OBB.



%
%

In this paper, to address the first problem, a stacking ensemble method is proposed. The stacking model is trained to best combine the member models, while simultaneously considering three factors, model calibration, model redundancy, and the performance gap between the models. For the second problem, a new bounding box fusion method is proposed for the oriented bounding boxes. The bounding boxes are parameterized with orientation, position, width, and height, and each parameter is fused separately. The combined method, OBBStacking, helps take 1st place in the Challenge Track \textit{Fine-grained Object Recognition in High-Resolution Optical Images}, which was featured in \textit{2021 Gaofen Challenge on Automated High-Resolution Earth Observation Image Interpretation}.

This paper is structured as follows. Related work will be discussed in Section \ref{sec:related_work}. The proposed ensemble method is introduced in Section \ref{sec:methods}. The experiment setup and the quantitative results are described in Section \ref{sec:results}. We also provide some analysis of OBBStacking in Section \ref{sec:discussion}. The conclusion is given at the end. 

\section{Related Work}\label{sec:related_work}
\subsection{Remote sensing object detection}

Quite a few deep neural network detectors are proposed in recent years. Notably, Liu et al. \cite{liu_rotated_2017} are among the earliest to utilize oriented bounding boxes (OBB) for object detection in remote sensing images. The method is built upon Faster RCNN \cite{ren_faster_2017} and proposes a rotated region of interest (RROI) pooling layer for accurate feature extraction; and an OBB regression model for precise object positioning. Later methods \cite{zhang_toward_2018, yang_automatic_2018, azimi_towards_2018} adopt oriented anchors for a better formulation of the bounding box that's easier to learn for the neural networks, but at the cost of relying on a redundant number of rotated anchors. Ding et al. \cite{ding_learning_2019} propose ROI Transformer to alleviate the problem by formulating RROI as offset parameters relative to only non-oriented ROIs. Han et al. \cite{han_redet_2021} build upon general rotation equivariant CNNs \cite{weiler_general_2019} and ROI Transformer to create an oriented object detection model (ReDet) with rotation equivariant features. Xie et al. \cite{xie_oriented_2021} further simplify the OBB inference process of ROI Transformer with 1/3000 number of parameters used and propose a new model, Oriented R-CNN, which is currently state-of-the-art on Dota \cite{xia_dota_2018} Dataset. 

ReDet and Oriented R-CNN are two of the models we select to generate the detection results for our ensemble method. This is due to their recognized performance on similar problems and their large backbone network difference, where ReDet uses rotation equivariant CNN and Oriented R-CNN uses the more traditional ResNet \cite{he_deep_2015} architecture. The intrinsic difference in their backbone will help increase the model diversity and in turn, increase the effectiveness of the ensemble process.

\subsection{Transformer}

Transformer is another neural network structure we take interest in, due to its structural difference from CNN. It was first introduced by Vaswani et al. \cite{vaswani_attention_2017} for the natural language processing (NLP) problem. It is designed for sequential data and is effective at modeling long-distance dependencies, which is typical in language data. Its success motivated its adaptation to the computer vision domain, with the major hurdle being the difference in the structuring of data (one dimension vs. two/three dimensions) and the increased data length at each dimension.

ViT \cite{dosovitskiy_image_2020} by Dosovitskiy et al. was one of the notable Transformer models for computer vision problems. ViT divides one full image into several small patches to be treated as tokens, like the words in NLP, and proposes large-scale pre-training to compensate Transformer's lack of intrinsic properties for image data, such as translation equivariance and feature locality.

Swin Transformer \cite{liu_swin_2021} is one of the latest vision Transformer models. Swin Transformer proposes to boost its efficiency by utilizing the locality characteristic of the images and increasing the scale of features step-by-step through a hierarchical design. Swin Transformer will also be one of the backbones for our member neural network detectors.

\subsection{Calibration of the neural networks}

A well-calibrated model can produce the probability of correctness for each prediction. Guo et al. \cite{guo_calibration_2017} show that while modern neural networks excel at making correct predictions, their level of calibration degrades. This hinders the attempt to effectively combine different neural networks and their application in critical scenarios. Guo et al. propose to calibrate the models in a post-processing manner and train a simple parametric model (Temperature Scaling) \cite{platt_probabilistic_1999} to map the confidence scores of the models to the probabilities of correctness. Wenger et al. \cite{wenger_non-parametric_2020} propose a latent Gaussian process to correct the model output. Zhang et al. \cite{zhang_mix-n-match_2020} propose an ensemble of post-processing methods that is data efficient and with high generalizability.

The above methods are post-processing calibration methods that are most related to our work. There are also calibration methods such as Bayesian neural network methods \cite{izmailov_subspace_2020} and neural network regulation methods \cite{pereyra_regularizing_2017-1} that change the design philosophy or the objective functions to achieve more calibrated neural networks.

\subsection{Bounding box post-processing methods}

Object detection methods, along with other vision-related algorithms, may produce redundant activations in a close spatial neighborhood. Non-Maximum Suppression (NMS) has been used in such scenarios for over half a century \cite{rosenfeld_edge_1971} and to this day, is still being used in the deep neural network pipelines. Specifically, modern neural network detectors generate redundant results for a single object and NMS post-processes the results by checking the spatial overlaps of the results and keeping the ones with the highest confidence scores.

NMS eliminates the redundant bounding boxes completely, which may lead to false negatives when there are overlaps between the ground truth bounding boxes. Soft-NMS \cite{bodla_soft-nmsimproving_2017} alleviates the problem by keeping all the bounding boxes and only mapping the confidence scores of the to-be-suppressed bounding boxes to a lower value.

Weighted Boxes Fusion (WBF) \cite{solovyev_weighted_2021} targets specifically at post-processing the bounding boxes from different models. Instead of selecting one best bounding box (NMS) or keeping all of the bounding boxes (Soft-NMS), WBF produces a weighted average of the bounding boxes in terms of position and size, so all of the to-be-fused bounding boxes can contribute to the final bounding box and no redundant bounding boxes are introduced.

\section{Methods}\label{sec:methods}

\begin{figure*}[t]
    \centering
    \includegraphics[width=\textwidth]{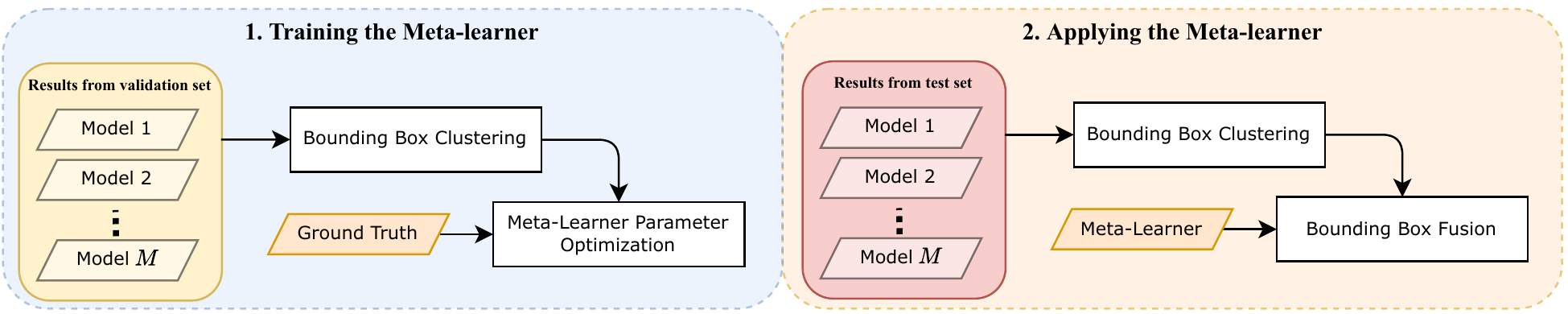}
    \caption{Two stages of the proposed OBBStacking}\label{fig:diagram}
\end{figure*}

OBBStacking is a stacking ensemble method that is compatible with OBBs. In a stacking method, a new model called a meta-learner, is trained to best combine the results of multiple existing models. OBBStacking has two stages (Fig. \ref{fig:diagram}), training the meta-learner, and applying the meta-learner to the member models. First, we will introduce the meta-learner proposed in our method. Then, we will be discussing the key processes that constitute the two stages---namely bounding box clustering, meta-learner parameter optimization and bounding box fusion.

\subsection{The Meta-Learner}
In a stacking method, every member model makes an independent prediction based on a data sample, and the meta-learner combines the predictions to form a more accurate one. In OBBStacking, we choose a simple model, logistic regression, as the meta-learner. The model takes the form

\begin{equation}
    \sigma_{\text{WA}}(\mathbf{z})=\sigma(\mathbf{zw}+b) \label{eq:meta-learner}
\end{equation}
where $\mathbf{z}=[z_1, z_2, ..., z_M] \in \mathcal{R}^{2 \times M}$ is the concatenation of the logit output from $M$ member models, $\sigma(z)=\frac{1}{1+\mathrm{exp}(-z)}$ is the logistic function, and $\mathbf{w} \in \mathcal{R}^{M}$ and $b \in \mathcal{R}$ are the weight and the intercept parameter of the meta-learner, respectively. 

Note that logit $z\in\mathcal{R}^2$ is the non-probabilistic output of the member models and the 2 dimensions correspond to the tendency of refusing a target and accepting a target, respectively. In the context of deep learning, logits are often converted to probabilistic output through the logistic function, but here the logits are used because of their amenity to Eq. \ref{eq:meta-learner}. 

Later in Section \ref{sec:discussion}, we will show how this simple form of the meta-learner can simultaneously consider model calibration, model redundancy and the performance gap between the models.

\subsection{Bounding box clustering}

Under the OBB detection setting, each member model produces a set of OBBs, but the correspondence of OBBs between different sets is unknown. Therefore the first goal is to collect output $\mathbf{z}$ on the same object from the different member models. We assume the OBBs are relatively accurate in terms of position and shape such that OBBs generated from the same object but different models have a significant spatial overlap. Therefore, an OBB spatial clustering method is used to assign OBBs from the same object into the same cluster.

The clustering method has the following steps:

\begin{enumerate}
    \item Aggregates all the OBBs from the member models into a list $\mathbf{S}$, sorted by their bounding box scores $s$ in descending order.
    \item Create an empty list $\mathbf{C}$ for the resulting clusters.
    \item Pop the first OBB from $\mathbf{S}$ as a new cluster center, and push the cluster into $\mathbf{C}$.
    \item Iterate through $\mathbf{S}$ and find OBBs from other member models and have an overlap greater than $iou_{\mathrm{thresh}}$ with the cluster center and move them from $\mathbf{S}$ to the new cluster.
    \item Go back to Step 3 and repeat until $\mathbf{S}$ is empty.
\end{enumerate}

Note that although both stages of OBBStacking include bounding box clustering, the method is applied to different sets of data. The whole scheme requires three sets of data, the training set, the validation set, and the test set. \textit{Training set} is used to train the member models. \textit{Validation set} is used to train the meta-learner (Stage 1 of OBBStacking). \textit{Testing set} is used for measuring the final performance of OBBStacking (Stage 2). Member models and the meta-learner are trained on separate data sets to prevent the meta-learner from favoring the member models that overfit the training set.

\subsection{Meta-learner Parameter Optimization}\label{sec:inference}




After the member models are trained on the \textit{training set} and produce $M$ sets of detection OBBs on the \textit{validation set}, the bounding box clustering method is applied to acquire the clustered OBBs $\mathbf{C}_{\mathrm{val}}=\{\mathbf{c}_i|i=1,2,...n\}$. Each OBB in a cluster $\mathbf{c}_i$ represents the prediction of a member model from one data sample $\mathbf{x}_i$. 

Here, the major role of the meta-learner is to fuse the bounding box scores $s$ in the same clusters. Note that we use the logit output $z$ in Eq. \ref{eq:meta-learner}. In most detectors, $s$ and $z$ can be acquired by keeping both outputs before and after the last logistic function. Additionally, in most clusters, one or more member models will be absent when they predict the probability is lower than a threshold. We set $z$ for these cases to a fixed negative value to keep the optimization simple.

We use Negative Log Likelihood (NLL) as the objective function, which can be formulated as:

\begin{align}
    \mathcal{L}&=-\sum^n_{i=1}\text{log}(\sigma_{WA}(\mathbf{z}_i)^{(y_i)}) \\
    &=-\sum^n_{i=1}\text{log}(\sigma(\mathbf{z}_i\mathbf{w}+b)^{(y_i)}) \label{eq:nll2}
\end{align}
where $y_i$ is the ground truth label of each cluster. To determine $y_i$, we calculate IOU (Intersection over Union) between the cluster center OBB and all the ground-truth OBBs in the validation set. A cluster is marked as a true positive ($y=1$) if it has an overlapped ground-truth OBB, and a false positive ($y=0$) otherwise.

Eq. \ref{eq:nll2} is a convex function regarding to $\mathbf{w}$ and $b$, and can be easily optimized.



\subsection{Oriented bounding box fusion}\label{sec:fusion}

Before this step, the trained member models produce $M$ sets of OBBs from the \textit{test set}, which are then clustered into $C_{\mathrm{test}}$ with the bounding box clustering method.

This step aims to fuse the OBBs $\mathbf{O}=\{\mathbf{o}_1, ..., \mathbf{o}_K\}$ that belong to the same cluster into one OBB. We represent an OBB with a 7-tuple:
\begin{equation}\label{eq:obb}
    \mathbf{o}=(x, y, w, h, \theta, z, l)
\end{equation}
where $x, y, w, h, z$ represent the center coordinates on the x-y axis, width, height, and logit score, respectively. $l \in \{1,2,...,M\}$ is the index of its source model. Orientation $\theta \in [0, \pi)$ represents the angle between the longest axis of the bounding box and the x-axis.

The fusion process needs to derive the first 5 elements in $\mathbf{o}$ to acquire the final OBB, and these elements will be fused separately. With regard to the first 4 elements, the fusion process can be formulated as, 
\begin{equation}
    \mathbf{o}^{(j)}_\mathrm{fused}=\frac{\sum_{p=1}^n \mathbf{o}^{(j)}_p s^*_p}{\sum_{p=1}^n s^*_p}, j=1, 2, 3, 4
\end{equation}
where $j$ is the index of the element in $\mathbf{o}$, $p$ is the index of the OBB in the cluster, $\mathbf{o}_f$ is the fused OBB. $s^*$ is the calibrated score derived from OBB's logit score and the weight parameters in Eq. \ref{eq:meta-learner}:

\begin{equation}\label{eq:score}
    s^*_p=\sigma(z^{(1)}_{p} \mathbf{w}^{(l_p)}+b)
\end{equation}
$s^*$ acts like an improved weight for each OBB that addresses the output calibration and the redundancy in the member models.

Orientation parameter $\theta$ receives special treatment due to its cyclic property. First, the orientation of the bounding box with the largest score $s^*$ is designated as the major orientation $\theta_{\mathrm{MJ}}$ of the cluster. Then, the fused orientation is determined by averaging the relative orientations to $\theta_{\mathrm{MJ}}$:

\begin{equation}
    \theta_f=\frac{\sum_{p=1}^n r(\theta_p,\theta_{\mathrm{MJ}}) s^*_p}{\sum_{p=1}^n s^*_p} + \theta_{\mathrm{MJ}}
\end{equation}
where $r$ is a bivariate function that calculates the relative difference of two angles while considering their cyclic property:

\begin{equation}
    r(\theta_1, \theta_2) = 
    \begin{cases}

  \theta_1-\theta_2, & \mathrm{for}\ \mathrm{abs}(\theta_1-\theta_2) \le \frac{\pi}{2}\\
  \theta_1-\theta_2 + \pi, & \mathrm{for}\ \theta_1-\theta_2 < -\frac{\pi}{2}\\
  \theta_1-\theta_2 - \pi, & \mathrm{for}\ \theta_1-\theta_2 > \frac{\pi}{2}
    \end{cases}
\end{equation}
Note that here we assume $\theta\in [0, \pi)$ since we do not discriminate between the head and the tail of an OBB.

Lastly, the score of the fused bounding box is determined with Eq. \ref{eq:meta-learner} with the learned meta-learner.





\section{Results}\label{sec:results}
\subsection{Datasets}

Two datasets are used to validate our method, FAIR1M dataset\cite{noauthor_fair1m_2022} and DOTA dataset\cite{xia_dota_2018}. Both datasets have an evaluation server that evaluates the detection results on a test set of which the ground truth labels are not shared publicly. Both these evaluation servers adopt mean average precision (mAP) as the evaluation criteria, consistent with PASCAL VOC 2007 \cite{everingham_pascal_2010} and VOC 2012.

\begin{figure*}[t]
    \centering
    \includegraphics[width=.95\textwidth,trim={0px 0px 300px 0px},clip]{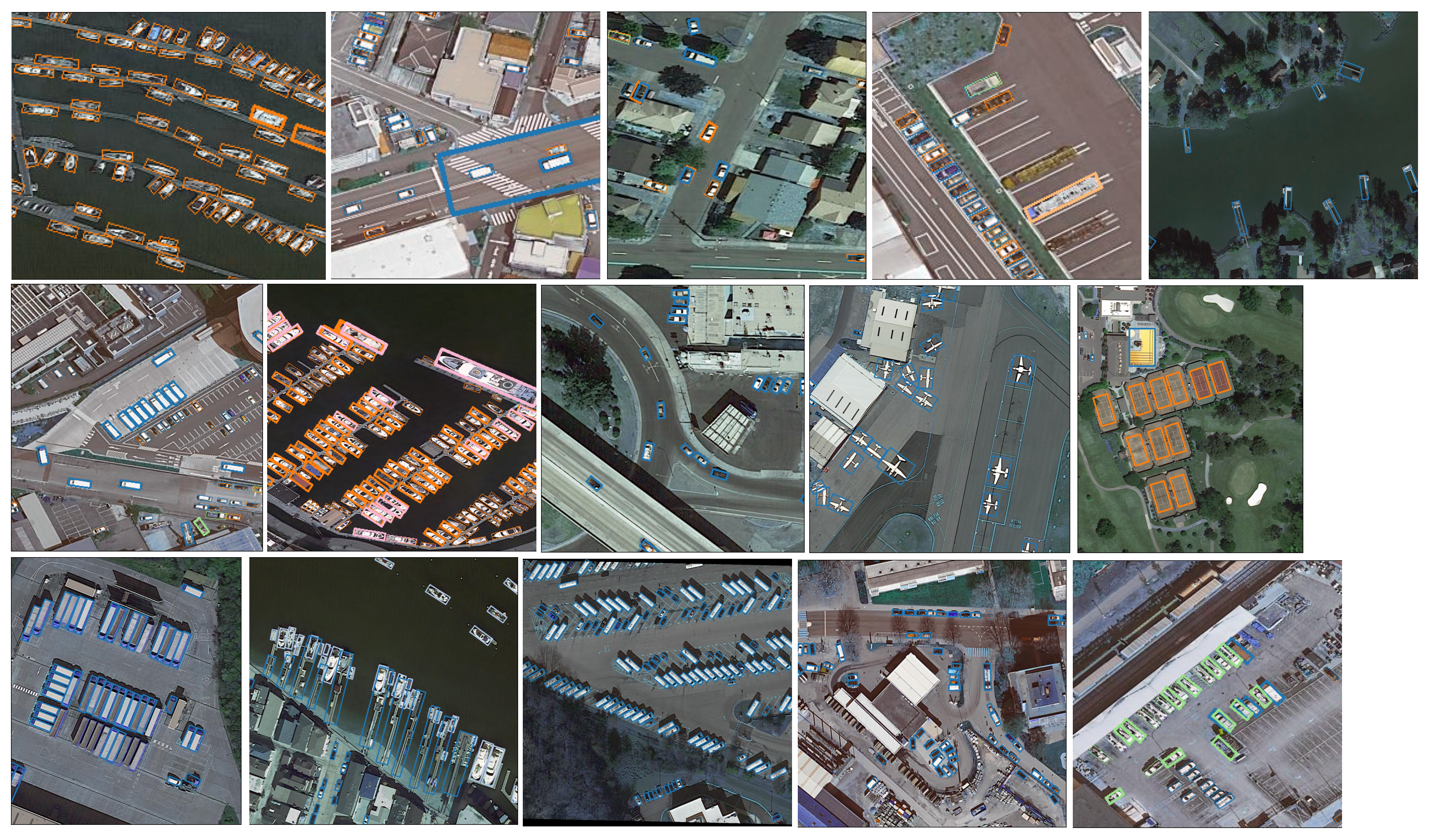}
    \caption{Showcase of the ensemble results of OBBStacking on DOTA dataset and FAIR1M dataset. Only the objects with a confidence score larger than 0.2 are shown.}\label{fig:showcase}
\end{figure*}

\subsubsection*{\bf FAIR1M dataset} This dataset was introduced alongside \textit{2021 Gaofen Challenge on Automated High-Resolution Earth Observation Image Interpretation}. It contains 32912 images with widths ranging from 600 pixels to 10000 pixels and spatial resolutions between 0.3 and 0.8 meters. The images are collected from Gaofen satellites and Google Earth, covering over 100 civil airports, harbors and cities. The dataset contains 1.02 million objects annotated with OBBs and assigned into 5 major categories and 37 fine-grained sub-categories. The major categories include vehicles, ships, airplanes, sports fields and road structures. The training, validation, and testing sets contain 16488, 8287, and 8137 images, respectively.

\subsubsection*{\bf DOTA dataset} This dataset was released in 2018. It contains 2806 images from satellites (GF-2 and JL-1), Google Earth, and aerial images with spatial resolution between 0.1 and 4.5 meters. It covers similar types of objects as FAIR1M does but with fewer sub-categories. It contains 15 categories and 0.2 million instances. The proportions of the training set, validation set, and testing set are 1/2, 1/6, and 1/3, respectively.

\subsection{Member Models}

As previously mentioned in Sec. \ref{sec:related_work}, we select 3 types of neural network detectors as the member models in the ensemble process, Oriented R-CNN, ReDet and a Swin detector. These 3 types of detectors have different design preferences so the diversity between the member models is assured.

The Swin detector in our experiment is a simple modification to the original one \cite{liu_swin_2021} for its compatibility with OBB detection. Both Swin backbone and the recent CNN backbone produce a feature pyramid \cite{lin_feature_2017}, consisting of layers of image features with different spatial resolutions and semantic depths, so their outputs have a similar structure and they can share the same types of detectors. We keep the original backbone and replace the original detector head with the one from Oriented R-CNN, since its OBB detector structure is elegant and concise.

For Oriented R-CNN and ReDet, we follow the experiment setups in the original papers, except for those that can be limited by the GPU specifications. We use a similar setting in Swin Detector to the ones in Oriented R-CNN since they share the same type of detectors. We use 2 GTX 3080 Ti for training and inference. The images are cropped into $1024 \times 1024$ patches and the batch size is set to 2, 2, and 1 per GPU for Oriented R-CNN, ReDet and Swin Detector, respectively, due to the limit of GPU memory. Multi-scale training and testing are also used because they are often used in combination with ensemble methods to achieve the highest performance possible.

\subsection{Quantitative Comparison}

First, for a fair comparison, we augment the original NMS and WBF with OBB compatibility, and evaluate the performance of the member models and the selected ensemble methods on DOTA dataset. Since most of the experiments in the literature \cite{xie_oriented_2021, han_redet_2021} combine the training and the validation sets to train their models to achieve maximum performance, and our ensemble model needs a separate validation set to learn the parameters of the meta-learner, we do two separate experiments to verify the effectiveness of our method. (1) We follow the original scheme of our method, and train all the member models on the training set only, leaving the validation set for the parameter training of the meta-learner. (2) We follow the training scheme of other methods and train the member models with data from both the training set and the validation set, and use the trained meta-learner from Experiment (1).

In the following tables on DOTA dataset, the names of the categories are abbreviated to conserve space. The categories, in order, are plane, baseball-diamond, bridge, ground-track-field, small-vehicle, large-vehicle, ship, tennis-court, basketball-court, storage-tank, soccer-ball-field, roundabout, harbor, swimming-pool, and helicopter.

\renewcommand{\arraystretch}{1.2}

\begin{table*}[t]
    \centering
    \caption{Quantitative results on Dota dataset, trained with the training set only}
    \label{tab:dota_train}
    \setlength{\tabcolsep}{0.8\tabcolsep}
    \begin{tabular}{l|lllllllllllllll|l} 
    
    \hline
    Methods & PL    & BD    & BR    & GTF   & SV    & LV    & SH    & TC    & BC    & ST    & SBF   & RA    & HA    & SP    & HC    & mAP    \\ 
    \hline
    \textbf{Individual} & & & & & & & & & & & & & & & &  \\
    Oriented R-CNN & 89.84 & 85.16 & 60.99 & 79.57 & 79.75 & 84.92 & 88.44 & 90.88 & 84.43 & 87.56 & 70.39 & 68.38 & 81.51 & 77.81 & 68.35 & 79.86  \\
    ReDet & 88.20 & 84.25 & 56.05 & 79.95 & 76.97 & 85.82 & 88.39 & 90.90 & 87.39 & 86.24 & 67.27 & 63.32 & 77.68 & 74.89 & 71.12 & 78.56  \\
    Swin Det & 88.77 & 81.99 & 57.59 & 76.63 & 65.26 & 84.24 & 87.96 & 90.83 & 84.49 & 87.24 & 63.36 & 66.45 & 80.74 & 67.34 & 65.82 & 76.58  \\ 
    \hline
    \textbf{Ensemble}   & & & & & & & & & & & & & & & &  \\
    NMS & 89.49 & 84.84 & 60.18 & 80.94 & 78.91 & 86.26 & 88.90 & 90.90 & 87.43 & 87.59 & 72.93 & 69.35 & 82.12 & 77.34 & 75.24 & 80.83  \\
    WBF & 89.49 & 84.94 & 60.20 & 80.94 & 78.99 & 86.25 & 88.90 & 90.90 & 87.43 & 87.84 & 73.06 & 70.62 & 82.45 & 76.13 & 75.24 & 80.89  \\
    Ours & 89.31 & 85.66 & 61.76 & 81.47 & 79.29 & 86.45 & 88.87 & 90.89 & 87.68 & 88.50 & 73.02 & 72.47 & 83.06 & 78.49 & 75.53 & 81.50  \\
    \hline
    \end{tabular}
\end{table*}

\begin{table*}[!t]
    \centering
    \caption{Quantitative Results on Dota dataset, trained with the training and the validation set combined}
    \label{tab:dota_trainval}
    \setlength{\tabcolsep}{0.8\tabcolsep}
    \begin{tabular}{l|lllllllllllllll|l} 
    \hline
    Methods       & PL    & BD    & BR    & GTF   & SV    & LV    & SH    & TC    & BC    & ST    & SBF   & RA    & HA    & SP    & HC    & mAP    \\ 
    \hline
    \textbf{Individual} & & & & & & & & & & & & & & & &  \\
    Oriented R-CNN      & 89.95 & 85.05 & 60.50 & 81.06 & 80.10 & 85.69 & 88.59 & 90.90 & 87.09 & 88.03 & 71.53 & 72.18 & 81.41 & 79.37 & 70.72 & 80.81  \\
    ReDet   & 88.28 & 84.82 & 59.13 & 78.56 & 77.23 & 85.83 & 88.71 & 90.88 & 87.17 & 86.75 & 67.31 & 65.79 & 78.23 & 78.82 & 69.85 & 79.16  \\
    Swin Det      & 89.66 & 83.79 & 59.39 & 76.22 & 76.57 & 84.15 & 88.49 & 90.87 & 83.61 & 86.59 & 61.95 & 62.00 & 80.79 & 69.93 & 72.62 & 77.77  \\ 
    \hline
    \textbf{Ensemble}   & & & & & & & & & & & & & & & &  \\
    NMS   & 89.62 & 85.21 & 61.05 & 78.88 & 79.73 & 86.52 & 89.05 & 90.90 & 86.59 & 87.64 & 72.07 & 68.35 & 82.84 & 79.81 & 76.12 & 80.96  \\
    WBF   & 89.62 & 85.57 & 60.91 & 78.88 & 79.88 & 86.53 & 89.06 & 90.90 & 86.59 & 87.90 & 72.07 & 72.42 & 83.13 & 80.03 & 75.94 & 81.30  \\
    Ours    & 89.68 & 85.79 & 62.52 & 80.32 & 80.10 & 86.75 & 89.05 & 90.86 & 87.38 & 88.26 & 72.79 & 72.25 & 83.89 & 79.74 & 76.68 & 81.74  \\
    \hline
    \end{tabular}
\end{table*}

\begin{table*}[!t]
    \centering
    \caption{Quantitative results on FAIR1M dataset}
    \label{tab:fair_train}
\begin{tabular}{l|l|ccc|ccc} 
    \hline
    \multicolumn{2}{l|}{\multirow{2}{*}{Methods}} & \multicolumn{3}{l|}{\textbf{Individual}} & \multicolumn{3}{l}{\textbf{Ensemble}} \\
    \multicolumn{2}{l|}{} & \multicolumn{1}{l}{Oriented R-CNN} & \multicolumn{1}{l}{ReDet} & \multicolumn{1}{l|}{Swin} & \multicolumn{1}{l}{NMS} & \multicolumn{1}{l}{WBF} & \multicolumn{1}{l}{Ours} \\ 
    \hline
    \multicolumn{1}{l}{} & mAP & 47.77 & 46.98 & 47.00 & 51.85 & 51.96 & 52.42 \\ 
    \hline
    \multirow{10}{*}{Plane} & Boeing737 & 47.95 & 43.54 & 36.60 & 51.38 & 51.38 & 51.60 \\
        & Boeing747 & 86.49 & 88.06 & 84.36 & 88.23 & 88.23 & 88.88 \\
        & Boeing777 & 30.61 & 25.89 & 21.46 & 34.97 & 34.97 & 34.27 \\
        & Boeing787 & 53.84 & 49.42 & 55.36 & 60.77 & 60.77 & 62.13 \\
        & C919 & 23.00 & 21.56 & 23.39 & 26.92 & 26.92 & 28.08 \\
        & A220 & 51.45 & 47.35 & 50.23 & 54.84 & 54.84 & 55.70 \\
        & A321 & 72.66 & 67.59 & 66.44 & 73.88 & 73.88 & 74.57 \\
        & A330 & 71.69 & 71.94 & 71.48 & 77.48 & 77.48 & 77.92 \\
        & A350 & 80.10 & 79.45 & 76.08 & 81.33 & 81.33 & 81.89 \\
        & ARJ21 & 41.40 & 44.58 & 35.70 & 46.60 & 46.60 & 48.67 \\ 
    \hline
    \multirow{8}{*}{Ship} & Passenger Ship & 16.62 & 22.42 & 19.34 & 23.38 & 23.38 & 24.19 \\
        & Motorboat & 68.83 & 74.73 & 71.56 & 75.81 & 75.91 & 76.48 \\
        & Fishing Boat & 12.68 & 15.77 & 10.95 & 16.19 & 16.27 & 15.83 \\
        & Tugboat & 29.67 & 40.03 & 38.01 & 41.54 & 41.37 & 42.72 \\
        & Engineering Ship & 15.72 & 16.28 & 19.41 & 20.85 & 20.85 & 21.03 \\
        & Liquid Cargo Ship & 31.14 & 30.24 & 29.90 & 35.69 & 35.81 & 35.69 \\
        & Dry Cargo Ship & 41.55 & 44.26 & 37.39 & 46.39 & 46.41 & 47.18 \\
        & Warship & 36.63 & 40.05 & 38.64 & 47.01 & 47.08 & 46.31 \\ 
    \hline
    \multirow{9}{*}{Vehicle} & Small Car & 77.47 & 71.84 & 73.39 & 76.59 & 77.32 & 77.60 \\
        & Bus & 56.06 & 44.26 & 55.43 & 59.53 & 59.54 & 59.97 \\
        & Cargo Truck & 55.30 & 49.26 & 55.18 & 57.89 & 58.11 & 58.67 \\
        & Dump Truck & 61.96 & 57.79 & 59.14 & 64.40 & 64.52 & 64.60 \\
        & Van & 77.66 & 72.57 & 73.96 & 75.73 & 76.05 & 76.23 \\
        & Trailer & 22.53 & 20.52 & 20.72 & 28.84 & 28.90 & 30.30 \\
        & Tractor & 7.82 & 3.61 & 6.47 & 7.55 & 7.55 & 8.10 \\
        & Excavator & 26.08 & 18.01 & 25.40 & 29.24 & 29.69 & 30.84 \\
        & Truck Tractor & 3.72 & 2.05 & 8.31 & 6.83 & 6.83 & 6.67 \\ 
    \hline
    \multirow{4}{*}{Court} & Basketball Court & 61.38 & 56.00 & 60.41 & 62.73 & 63.18 & 63.33 \\
        & Tennis Court & 88.11 & 87.80 & 86.76 & 90.21 & 90.56 & 90.36 \\
        & Football Field & 64.86 & 72.42 & 71.02 & 72.71 & 73.24 & 74.12 \\
        & Baseball Field & 89.11 & 90.02 & 88.85 & 91.45 & 91.45 & 91.40 \\ 
    \hline
    \multirow{3}{*}{Road} & Intersection & 62.20 & 62.83 & 63.74 & 64.71 & 64.80 & 65.26 \\
        & Roundabout & 27.48 & 17.59 & 18.76 & 28.89 & 28.89 & 27.48 \\
        & Bridge & 30.72 & 47.66 & 44.05 & 42.40 & 42.49 & 44.06 \\
    \hline
    \end{tabular}
\end{table*}

The quantitative results of Experiment (1) are listed in Table \ref{tab:dota_train}.  Oriented R-CNN achieves the best performance among the member models and obtains 79.86\% mAP. The ensemble methods all obtain a 1-2\% mAP increase over the best member model and our method achieves the top score with 81.50\% mAP, 0.61\% over WBF.

For Experiment (2), we assume the performance gap, the calibration, and the redundancy of the member models do not drift too much from Experiment (1), and we could reuse the meta-learner for the ensemble. The results are shown in Table \ref{tab:dota_trainval}. The results are generally similar to the previous one, with a slight overall performance increase of 1\% mAP among the member models and 0.1-0.4\% mAP increase among the ensemble methods. Our method, with the meta-learner from Experiment (1), still outperforms WBF by 0.24\% mAP. This shows that our assumption holds when the training data expands, and even though our method requires a separate validation set, it still outperforms the existing ensemble methods.

Next, we evaluate the member models and the ensemble models on FAIR1M dataset using Experiment (1) setup and show the results in Table \ref{tab:fair_train}. Among the member models, Oriented R-CNN still achieves the best performance with 47.77\% mAP. Compared to the individual methods, the ensemble models obtain a huge performance increase by around 4\% mAP, where our method achieves the best score with 52.42\% mAP, a 4.65\% increase over Oriented R-CNN, a 0.57\% mAP increase over WBF.

\section{Discussion}\label{sec:discussion}

In this section, we demonstrate how OBBStacking addresses the three problems that arise during an ensemble process on deep learning models---namely model calibration, the performance gap between the models, and model redundancy.

\subsection{Model Calibration}

Deep learning models tend to overfit the training data and are overconfident about their predictions. When the member models are overconfident to different degrees, their predictions are on different measurements and do not indicate true probability values. Therefore, the ensemble methods may not work well on these models as intended, and a model calibration process is needed.


In this section, we show that one of the calibration methods, Temperature Scaling (TS) \cite{guo_calibration_2017}, can be regarded as a special form of our meta-learner, indicating that OBBStacking includes the feature of model calibration.

TS attempts to map the non-accurate predictions to the real probability of correctness, by 'softening' the final logistic layer in the neural networks and introducing a \textit{temperature} parameter $T > 1$. The 'softened' logistic layer is
\begin{align}
    \sigma_{\text{TS}}(z)& =\sigma(z/T+t) \\ \label{eq:ts}
                         & =\frac{1}{1+\mathrm{exp}(-z/T+t)}
\end{align}
When $T\to \infty$, all results of $\sigma_{\text{TS}}$ approach $\frac{1}{2}$ and indicate maximum uncertainty.

The inference of parameter $T$ also uses NLL as the objective function, since NLL is a standard measure of a probabilistic model's quality \cite{hastie_elements_2009}. Here, the objective function can be defined as:

\begin{equation}\label{eq:nll}
    \mathcal{L}=-\sum^n_{i=1}\text{log}(
        \sigma(z_i/T+t)^{(y_i)}
    )
\end{equation}

As can be seen, our meta-learner, Eq. \ref{eq:meta-learner} becomes Eq. \ref{eq:ts} when the number of the member models is 1 and thus can calibrate models in the same fashion.


\subsection{Performance Gap}


In this section, we experiment to try to demonstrate how OBBStacking adjusts the weights when there is a performance gap between the models.

Our model tackles three problems simultaneously, model calibration, redundancy and performance gap. We assume these three problems can be disentangled and thus the factorization of the parameter exists, $\mathbf{w}=\mathbf{p}\odot \mathbf{r}\odot \mathbf{g}$, where the operator $\odot$ is the elementwise multiplication, $\mathbf{p, r, g}$ are the weight vectors for the model calibration, model redundancy and the performance gap, respectively.

We want to minimize the effect of the first two factors and see how OBBStacking handles the performance gap between the models. Along with the Swin detector used in our previous experiment, 3 additional Swin detectors are added to the Swin detector family. The only difference between these Swin detectors is the total number of epochs used in training, which are 12, 9, 16, and 18 epochs, respectively. At different epochs during the training with stochastic gradient descent, the neural networks may randomly lean towards more accuracy on some categories instead of others, and rely upon different features, thus creating a sequence of different models with relatively high redundancy.

\begin{figure}[t]
    \centering
    \includegraphics[width=.4\textwidth,trim={40px 20px 50px 30px},clip]{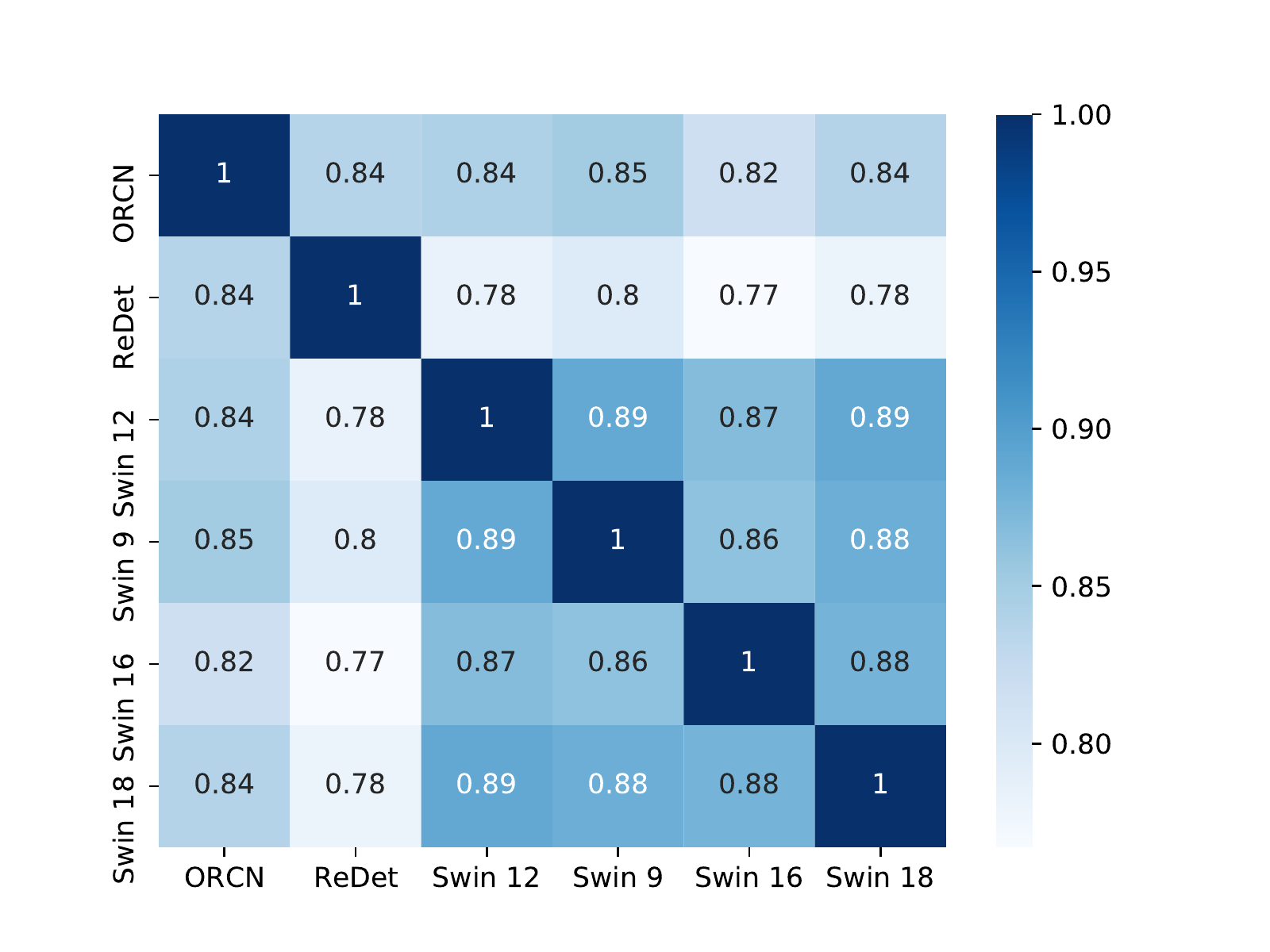}
    \caption{The correlation coefficient of models in Collection 2.}\label{fig:correlation2}
\end{figure}

We first run OBBStacking on the Swin family and acquire $\mathbf{w}$ for later comparison. Then, to show the factor of redundancy among the Swin family, we apply the bounding box clustering method to the detection results and calculate Pearson's correlation between the confidence scores of the models. As can be seen in Fig. \ref{fig:correlation2}, compared to the other models, the correlation coefficient between the Swin models are very close to each other, so we assume $\mathbf{r}$ is approximate to a vector of 1s.

As for the weight vector $\mathbf{p}$ from model calibration, it can be easily derived by applying TS on the Swin detectors individually, and we get $\mathbf{p}=[\frac{1}{T_1}, \frac{1}{T_2}, ..., \frac{1}{T_M}]$.

We list the above results and the separate mAP performances of the models in Table \ref{tab:discuss1}. As can be seen, the weight factor $\mathbf{g}$ is correlated to the mAP performance of the individual models. Model Swin 9, the model trained to 9 epochs, has the best prediction mAP on the validation set and the largest value in $\mathbf{g}$. Swin 16 has the worst mAP and also the smallest value in $\mathbf{g}$. This is in accord with the basic ensemble idea of putting more weight on the better predictors. Swin 12 and Swin 18 have similar values in $\mathbf{g}$ and similar performance in mAP, which is a reasonable range considering the small performance gap between the two models and the error from the assumed $\mathbf{r}$ value.


\begin{table}
    \centering
    \caption{Weight vectors of the Swin family}
    \label{tab:discuss1}
    \begin{tabular}{l|llll} 
        \hline
        Models     & Swin 12 & Swin 9 & Swin 16 & Swin 18  \\ 
        \hline
        $\mathbf{w}$ & 0.1705  & 0.2062 & 0.1283  & 0.1542   \\
        $\mathbf{r}$ & 1       & 1      & 1       & 1        \\
        $\mathbf{p}$ & 0.5028  & 0.5690 & 0.4406  & 0.4504   \\
        $\mathbf{g}$ & 0.3390  & 0.3625 & 0.2912  & 0.3423   \\
        mAP          & 48.21   & 48.80  & 46.63   & 47.99    \\
        \hline
        \end{tabular}
\end{table}

\subsection{Model Redundancy}

In this part, we build upon the previous experiments to show how OBBStacking handles model redundancy. 2 collections of models are included. Collection 1 consists of Oriented R-CNN, ReDet and Swin 12. Collection 2 includes all the models in Collection 1 and the additional Swin 9, Swin 16 and Swin 18, adding up to 6 models in total.

The correlation coefficient between all the models is shown in Fig. \ref{fig:correlation2} and the weight parameters $\mathbf{w}$ of the meta-learner are shown in Table \ref{tab:discuss2}. We notice that in Collection 2, because of the redundancy among the Swin families, their weights decrease drastically, with a sum value of 0.36, in between the weights of Oriented R-CNN and ReDet. The weights of Oriented R-CNN and ReDet decrease slightly because the Swin family improves its performance with the increase of its members.

\begin{table}
    \centering
    \caption{Weight vectors of Collection 1 and Collection 2}
    \label{tab:discuss2}
    \begin{tabular}{c|cccccc} 
    \hline
    \textbf{Collections}   & OR-CNN & ReDet & Swin 12 & 9 & 16 & 18  \\ 
    \hline
    \textbf{1}    & 0.57   & 0.34  & 0.24    & -      & -       & -        \\
    \textbf{2}    & 0.45   & 0.24  & 0.09    & 0.10   & 0.08    & 0.09     \\
    \hline
    \end{tabular}
\end{table}

\section{Conclusion}

We propose an ensemble method, OBBStacking, that is compatible with the oriented bounding box (OBB) which is widely used in object detection in the remote sensing field. OBBStacking consists of a meta-learner that can address the problems in the ensemble process of the deep neural network detectors, namely the model calibration, the redundancy between the models and the performance gap between the models. OBBStacking outperforms other ensemble methods in the DOTA dataset and the FAIR1M dataset and helps us win 1st place in the Challenge Track \textit{Fine-grained Object Recognition in High-Resolution Optical Images} featured in \textit{2021 Gaofen Challenge on Automated High-Resolution Earth Observation Image Interpretation}.

\bibliographystyle{IEEEtran}
\bibliography{IEEEabrv,main.bbl}

\end{document}